\documentclass[runningheads]{llncs}

\usepackage{accv}
\usepackage{accvabbrv}
\usepackage{graphicx}
\usepackage{booktabs}
\usepackage{array} 
\usepackage{amssymb}
\usepackage[accsupp]{axessibility}  
\usepackage[pagebackref,breaklinks,colorlinks,citecolor=accvblue]{hyperref}
\usepackage{orcidlink}
\usepackage{authblk}

\makeatletter
\def\thanks#1{\protected@xdef\@thanks{\@thanks
        \protect\footnotetext{#1}}}
\makeatother

\begin{document}
    \title{SGW-based Multi-Task Learning in Vision Tasks}

    \author{
        Ruiyuan Zhang\inst{1,\star} \and
        Yuyao Chen\inst{1,\star}\thanks{$\star$~Equal contributions.} \and
        Yuchi Huo \inst{1} \and
        Jiaxiang Liu \inst{1} \and
        Dianbing Xi \inst{1} \and
        Jie Liu \inst{1} \and
        Chao Wu\inst{1,\dag} \thanks{\dag~Corresponding Author}
    }
    
    \institute{
        Zhejiang University, Hangzhou, Zhejiang, China \\
        \email{\{zhangruiyuan, zjljx, db.xi, chao.wu@\}zju.edu.cn} \\
        \email{huo.yuchi.sc@gmail.com} ~~
        \email{nothing207@yeah.net} ~~
        \email{liujiemath@hotmail.com} 
    }

    \maketitle

    \begin{abstract}
        Multi-task-learning(MTL) is a multi-target optimization task. Neural networks try to realize each target using a shared interpretative space within MTL. 
        However, as the scale of datasets expands and the complexity of tasks increases, knowledge sharing becomes increasingly challenging.
        In this paper, we first re-examine previous cross-attention MTL methods from the perspective of noise.
        We theoretically analyze this issue and identify it as a flaw in the cross-attention mechanism. To address this issue, we propose an information bottleneck knowledge extraction module (KEM). This module aims to reduce inter-task interference by constraining the flow of information, thereby reducing computational complexity.
        Furthermore, we have employed neural collapse to stabilize the knowledge-selection process. That is, before input to KEM, we projected the features into ETF space. This mapping makes our method more robust.
        We implemented and conducted comparative experiments with this method on multiple datasets. The results demonstrate that our approach significantly outperforms existing methods in multi-task learning. 
        
    \end{abstract}

    \section{Introduction}
        Some tasks are similar, such as segmentation and depth prediction. So, it's natural to consider multi-target optimization.
        Within multi-task learning, we can set a shared interpretative space for image\cite{xin2024mmap}. The key idea is to utilize the intrinsic correlation between tasks to share and transfer knowledge among different tasks. Under this paradigm, MTL has been successfully applied in many fields \cite{hu2024revisiting, chen2023minigpt, muhammad2020deep, wang2020graph, zhang2021survey}.
        A common pipeline for MTL is to design models using a modular approach, employing cross-attention mechanisms for knowledge sharing \cite{xu2022mtformer, han2022survey}. MTFormer\cite{xu2022mtformer} utilizes a shared encoder module to process inputs and then feeds this knowledge to each task-specific processing module. Inter-task attention is used to exchange information between tasks. MTFormer has already demonstrated its superior performance.

\begin{figure}[t]
            \centering
            \includegraphics[width=1.0\textwidth]{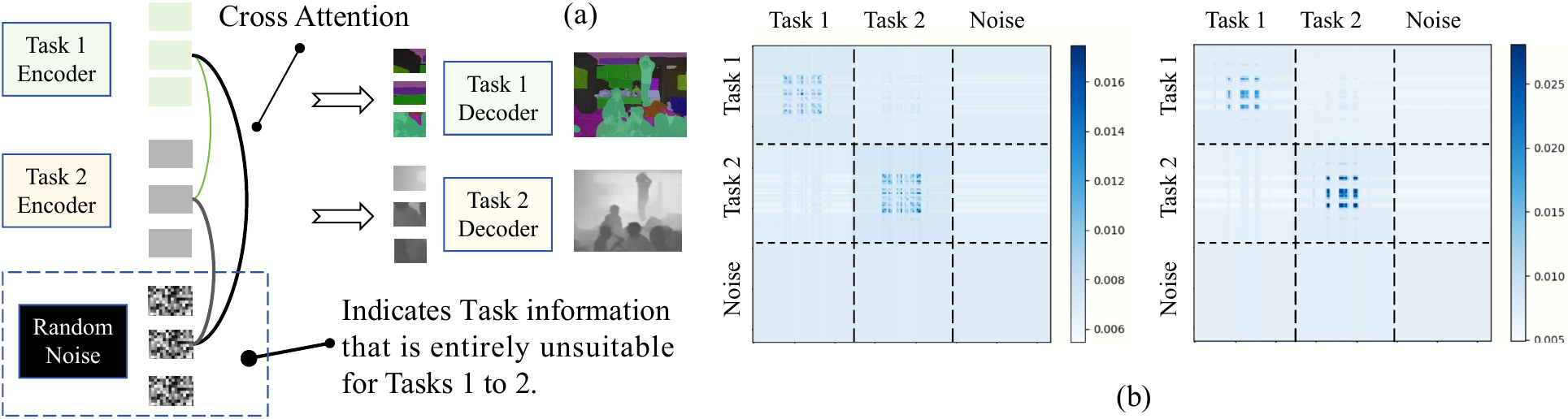}
            \caption{
                Toy experiment to illustrate the concept of noise. For simplicity, we utilized a 2-layer single-head attention mechanism and added standard Gaussian noise to the hidden states to simulate irrelevant information in Multi-Task Learning (MTL).
            }
            \label{fig:noise}
        \end{figure}

\begin{figure}[t]
            \centering
            \includegraphics[width=1.0\textwidth]{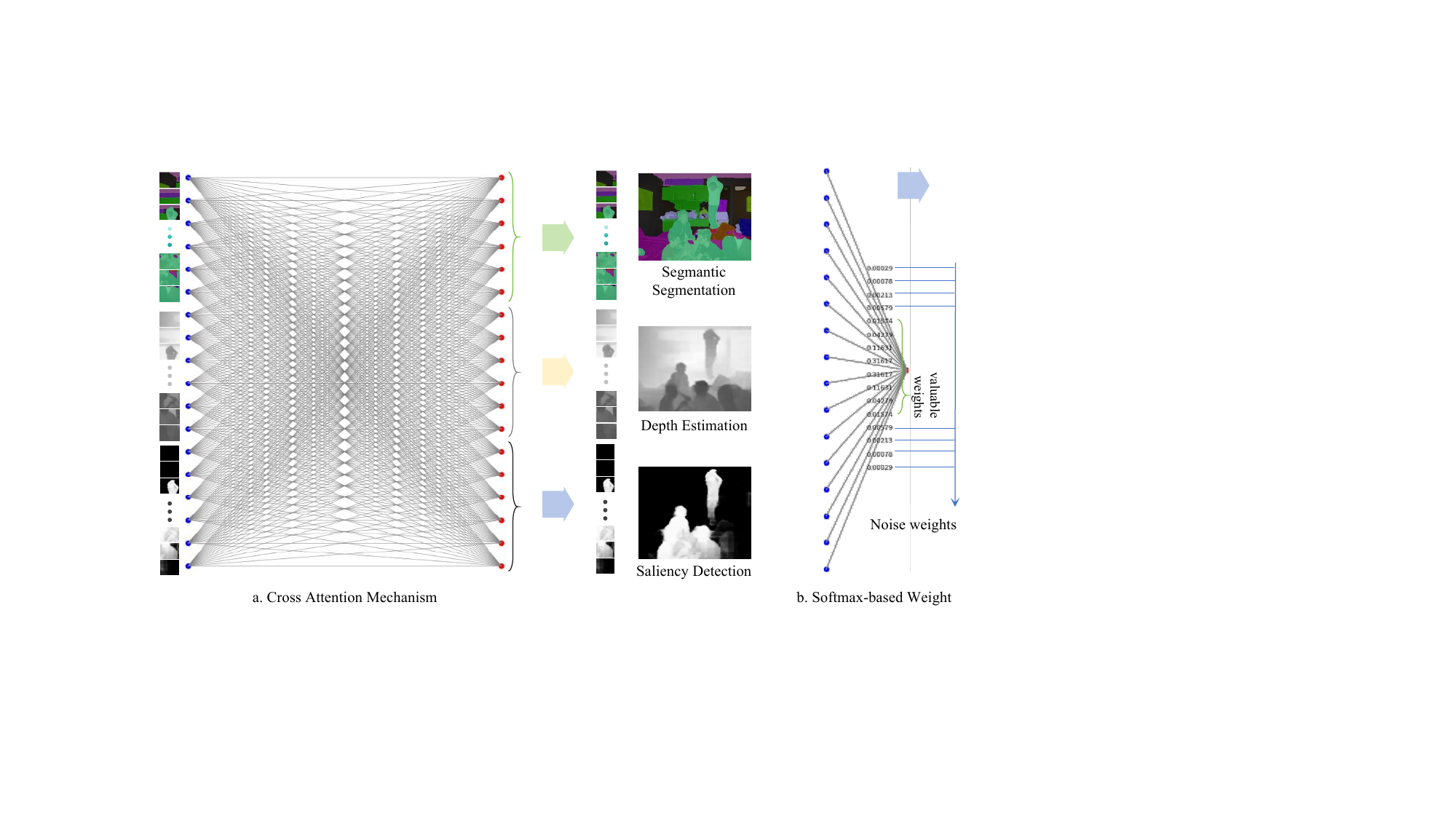}
            \caption{
                The illustration of Knowledge-extraction module in multi-task learning using Cross Attention Mechanism.
                (a) Knowledge extraction from the feature spaces of other tasks, showing $\mathcal{O}(2n_s^2 \cdot d_e)$ computational complexity.
                (b) Attention weights calculation using Softmax, which can be divided into noise and valuable weights, is one reason for inter-task interference.
            }
            \label{fig1}
        \end{figure}

        However, only a select few factors of variation are relevant for each downstream task. This becomes problematic as cross-attention mechanisms struggle to share knowledge within large-scale datasets or complex tasks\cite{lachapelle2023synergies, xu2022mtformer, liu2023parameter, liu2023deep}. 
        As shown in Fig.\ref{fig1}. a) As the number of tasks and their complexity increase, the computational demands grow quadratically. b) A significant issue arises with the softmax operation, which rarely results in zero weights for any position, leading to the creation of unnecessary \textbf{noise} that interferes with task performance. To clarify the motivation, we designed a toy experiment under extreme conditions.
        As illustrated in Figure~\ref{fig:noise}(a), we introduce Task 3, which contains entirely noise, to show that it can still influence Tasks 1 and 2 following the Softmax operation. This setup can verify that Softmax alone cannot effectively eliminate noise. The probabilities in real scenarios are visualized in Figure~\ref{fig:noise}(b). Tasks 1 and 2 successfully share knowledge, demonstrating that MTL facilitates knowledge transfer between task encoders. However, they still acquire approximately 20\% of explicit noise information. Therefore, we can infer that irrelevant information in Task 1 may be transformed into noise after the Softmax operation, subsequently influencing Task 2.

        Following this key insight, we propose a Multi-task Knowledge Extraction Module (KEM). KEM compresses the knowledge from tasks into an information-bottleneck memory and then distributes the memory to different tasks. 
        We divide this process into three steps: Retrieve, Write, and Broadcast. This is essentially a selection mechanism that utilizes an additional memory slot to filter out noise from the input features $F$, retaining only the useful data.
        Due to the choice mechanism, KEM can eliminate ineffective noise information, retain common information, and resist interference between tasks. 
        The memory size $L$ is constant, and the accompanying benefit is that we can also reduce the algorithmic complexity to a linear level.

        We adopted the Top-K choice and completed the write operation in memory, as shown in Equation~\ref{eq:step1}. However, the Top-K is easily influenced by the statistical distribution of input data \cite{yang2022inducing, shwartz2024simplifying, dang2024neural}.
        Additionally, the input feature distribution given to KEM is uncontrolled; therefore, we must mitigate the impact of unknown data distributions on the Top-K choice to generalize KEM to other tasks. 

        To further enhance the stability of the knowledge selection process, we introduce the concept of neural collapse \cite{tirer2023perturbation, zhong2023understanding, zhu2021geometric}, and propose Stable-KEM(sKEM).  Neural collapse is a phenomenon where, in the later stages of model training, the features of samples belonging to the same class converge to the same direction. This phenomenon helps stabilize the knowledge selection process, making our model more robust.
        sKEM projects the input features into an Equiangular Tight Frame(ETF space), allowing memory to select even statistical-less features. We have theoretically shown that sKEM can improve resistance within imbalanced input. We also confirmed sKEM's effectiveness through experiments. 

        In summary, we address inter-task interference in multi-task learning by proposing KEM and analyzing its computational complexity. We further enhance KEM's stability using Neural Collapse, leading to the development of sKEM, whose stability we have theoretically proven. Lastly, we validate the effectiveness of both sKEM and KEM through comparisons with other baselines.
        
    \section{Related Works}
        \subsection{Multi Task Learning}
            Multi-task learning (MTL) is an intriguing area of study. It uses a single model to predict multiple targets, such as semantic segmentation, human parts segmentation, depth estimation, surface normal estimation, and boundary detection. Technically, MTL frameworks can be categorized into CNN-based \cite{gao2019nddr, vandenhende2020mti}, NAS-based \cite{bhattacharjee2022mult}, and Transformer-based \cite{vandenhende2021multi} models. For instance, Xu \cite{xu2022mtformer} employed a pre-trained model to enhance MTL performance and implemented cross-attention mechanisms for knowledge sharing.
            In the line of disentangled representations, several studies \cite{dittadi2020transfer, miladinovic2019disentangled, montero2020role} have empirically explored this question and obtained positive results. Theoretically, Lachapelle \cite{lachapelle2023synergies} discovered that only a small subset of all factors of variations is useful for each downstream task. We support this assumption and view knowledge sharing in MTL from the perspective of distinguishing between valuable and noisy weights.

        \subsection{Information Bottleneck Module}
            The success of VQ-VAE\cite{van2017neural, razavi2019generating} has demonstrated that it is possible to represent high-dimensional data using low-dimensional discrete latent variables. This approach not only effectively captures the structure of the data but also removes noise through quantization processes\cite{im2017denoising}, thereby achieving higher resolution in the reconstruction process\cite{menon2020pulse}. 
            Based on this research, Tucker\cite{tucker2021emergent} applies discrete tokens in multi-agent communication, making agent performance robust to environmental noise. 
            Instead of Discrete IB, Liu\cite{liu2022stateful} employs a Soft IB named SAF, a shared knowledge source for sifting through and interpreting signals from all the agents before passing them to the centralized critic. Zhang\cite{zhang2024scalable} proposes a co-creation space among assemblers, solving scalability issue. Some works\cite{goyal2021coordination, hong2024concept} combined the Shared Global Workspace Theory to propose their understanding of soft IB, improving single task's performance.
            
        \subsection{Neural Collapse}
            Neural Collapse\cite{papyan2020prevalence} is a phenomenon that occurs during the training of a classification model on a balanced dataset. 
            During the later stages of training, the features in the last layer of a neural network will converge toward specific centers, creating an Equiangular Tight Frame(ETF) with clearly identifiable symmetric properties.
            Training the network along this phenomenon can achieve better performance \cite{zhong2023understanding, wang2024balance}.
            Later theoretical explorations\cite{fang2021exploring, ji2021unconstrained, zhu2021geometric, tirer2022extended} aimed to unravel the complexity of this phenomenon shows that neural networks after NC exhibit higher linear separability.
            Following those, some research\cite{yang2022inducing, shwartz2024simplifying, zhong2023understanding} has studied how to use NC under imbalance input.
            In this paper, we project the input features into ETF space as \cite{yang2022inducing, shwartz2024simplifying}, establishing a stable Top-K choice in KEM.

    \section{KEM: Multi-Task Knowledge Extraction Module} \label{sec:kem}
        As we said before, we view knowledge sharing in MTL from the perspective of value and noise weights. So, we propose an information bottleneck module named KEM in \ref{sec:kem:main}. Furthermore, we conducted a complexity analysis on KEM in \ref{sec:kem:complexity}.
        \begin{figure}[t]
            \centering
            \includegraphics[width=1.0\textwidth]{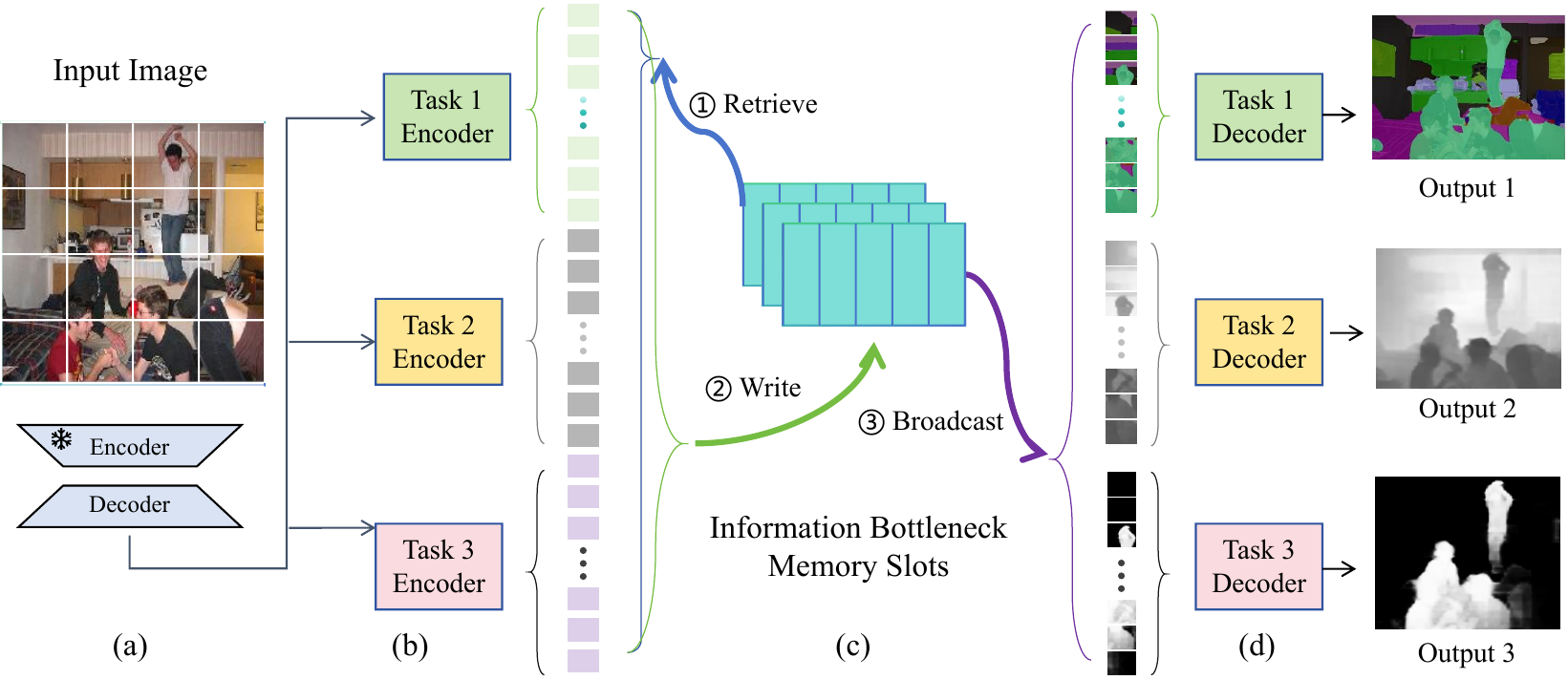}
            \caption{
                The illustration of the proposed KEM framework. 
                (a) A frozen pre-trained Vision Transformer model is used for initial image encoding.
                (b) Each task learns its encoder.
                (c) Features between tasks interact through an information bottleneck memory slots, involving three steps: Retrieve, Write, and Broadcast. See Section~\ref{sec:kem} for details.
                (d) Each task learns its decoder to decode data into the task-specific format.
            }
            \label{fig:framework}
        \end{figure}

        \subsection{Architecture Design} \label{sec:kem:main}
            As shown in Figure~\ref{fig:framework}, KEM comprises four components.
            The network processes RGB images as input. KEM uses a pre-trained Swin-Transformer model for initial image encoding, with the encoder part frozen and only the decoder learning during training. The image was split into \( m \) patches, each with a dimension of \( d_p \).
            To enhance model efficacy across diverse tasks, we have designed specialized encoders and decoders for each task, realized as a 2-layer Vision Transformer (ViT).
            We suppose there are $n$ tasks and thus corresponding $n$ task features. The total output features are represented as $F = F_1 \oplus F_2 \oplus \cdots \oplus F_n$, where $F_i \in \mathbb{R}^{m \times d_p}$ represents the features of the $i$-th task.
            After the task-specific encoder, the features will be extracted through a memory slot $R \in \mathbb{R}^{L \times d_l}$. $R$ consists of $L$ slots ${l_0, l_1, ..., l_{L-1}}$.
            There are three steps: Retrieve, Write, and Broadcast. 
            
            \textbf{Step 1: Retrieve the value of $F$} KEM uses the memory slots $R$ in order to update itself.
            Among them, $R$ is a learnable parameter to retrieve data within $F$.
            The memory slot $R$ is updated as:
            \begin{equation} \label{eq:step1}
                \hat{R} = \text{topk-softmax}(\frac{(RW^{qr})(FW^{kr})^T}{\sqrt{d_e}})FW^{vr},
            \end{equation}
            where $W^{qr}$, $W^{kr}$, $W^{vr}$ represent the projection matrices.
            
            \textbf{Step 2: Write $\hat{R}$ into each Task $\hat{F}$.}
            The knowledge in $\hat{R}$ is inconsistent with $F$ regarding dimensionality. Thus, we need to employ attention mechanism once more to write the knowledge from $R$ into $\hat{F}$, where the shape of $F$ equals $\hat{F}$.
            \begin{equation}\label{eq:step2}
                \hat{F} = \text{softmax}(\frac{(FW^{qw})(\hat{R}W^{kw})^T}{\sqrt{d_e}})\hat{R}W^{vw},
            \end{equation}
            where $W^{qw}$, $W^{kw}$, $W^{vw}$ represent the projection matrices.

            \textbf{Step 3: Broadcast $\hat{F}$ through residual connection.} $w_r$ is a weight for residual connection, we set $w_r$ = 1.
            \begin{equation} \label{eq:step3}
                F \leftarrow{} F + w_r \hat{F}. 
            \end{equation}
            
            Generally speaking, the knowledge extraction module acquires representations as inputs from various tasks and ultimately broadcasts the outputs to each task. 
            It is task-agnostic and can seamlessly replace the existing cross-attention mechanism without complications.
            
        \subsection{Complexity Analysis for KEM}\label{sec:kem:complexity}
           The formulation of cross-attention in Figure~\ref{fig1} is:
            \begin{equation}
                F' = \text{softmax}\left(\frac{(FW^{q})(FW^{k})^T}{\sqrt{d_e}}\right)FW^{v}.
            \end{equation}
            Here, $F$, and $F'$ are $n_s \times d_e$ matrices, where $n_s = n \times m$ is the sequence length and $d_e$ is the feature dimension. First, we calculate $FW^{q}$ and $FW^{k}$, each with a complexity of $\mathcal{O}(n_s \cdot d_e^2)$. Secondly, we compute $(FW^{q})(FW^{k})^T$, which is an $n_s \times n_s$ matrix multiplication with a complexity of $\mathcal{O}(n_s^2 \cdot d_e)$. And then, we calculate another matrix multiplication with $FW^{v}$, which also has a complexity of $\mathcal{O}(n_s^2 \cdot d_e)$. The computational complexity of the remaining parts can be similarly calculated. The overall computational complexity is approximately: 
            \[
                \mathcal{O}(2n_s^2 \cdot d_e + 3n_s \cdot d_e^2)+\mathcal{O}(n_s^2).
            \]
            
            KEM's computation can be split into Equation~\ref{eq:step1} and Equation~\ref{eq:step2}.
            In Equation~\ref{eq:step1}, $R$ is an $L \times d_e$ matrix, where $L$ is the length of the memory slots. Similarly, we calculate $RW^{qr}$ and $FW^{kr}$, with complexities of $\mathcal{O}(L \cdot d_e^2)$ and $\mathcal{O}(n_s \cdot d_e^2)$ respectively. Then we compute $(RW^{qr})(FW^{kr})^T$, an $L \times n_s$ matrix multiplication with a complexity of $\mathcal{O}(L \cdot n_s \cdot d_e)$. The remaining parts can be calculated analogously. The total computational complexity is:
            \[
                \mathcal{O}(2L \cdot n_s \cdot d_e + L \cdot d_e^2 + 2n_s \cdot d_e^2)+\mathcal{O}(L\cdot n_s).
            \]
            
            For Equation~\ref{eq:step2}, we calculate $FW^{qw}$ and $\hat{R}W^{kw}$, with complexities of $\mathcal{O}(n_s \cdot d_e^2)$ and $\mathcal{O}(L \cdot d_e^2)$ respectively. Then we compute $(FW^{qw})(\hat{R}W^{kw})^T$, with a complexity of $\mathcal{O}(n_s \cdot L \cdot d_e)$. And we calculate other parts similarly. The overall complexity for this equation is: 
            \[
                \mathcal{O}(2n_s \cdot L \cdot d_e + n_s \cdot d_e^2 + 2L \cdot d_e^2)+\mathcal{O}(L\cdot n_s).
            \]

            Combining the complexity analyses in the KEM method, the total computational complexity can be expressed as:
            \[
                \mathcal{O}(4L \cdot n_s \cdot d_e + 3n_s \cdot d_e^2 + 3L \cdot d_e^2)+\mathcal{O}(L\cdot n_s).
            \]

            Given that \( n_s > d_e > L \), the total computational complexity for cross-attention is $\mathcal{O}(2n_s^2 \cdot d_e).$ Meanwhile, the total computational complexity for KEM is $\mathcal{O}(3n_s \cdot d_e^2)$, which implies that KEM has a lower computational complexity. 

        \subsection{Training Loss}
            To optimize the multi-task learning framework, we employ a classical MTL loss function, which is designed to balance the training across multiple tasks effectively. The MTL loss function is a combination of individual loss functions for each task, weighted by a set of task-specific parameters that are optimized during training.
            The general form of the MTL loss function is given by:
            \begin{equation}
                L(\theta) = \sum_{i=1}^{n} \alpha_i L_i(\theta_i).
            \end{equation}
            where \(L(\theta)\) is the total loss for the model, \(L_i(\theta_i)\) represents the loss function for the \(i\)-th task, \(\theta_i\) are the parameters specific to the \(i\)-th task, and \(\alpha_i\) are the task-specific weights. 
            
            These weights are crucial as they help in prioritizing some tasks over others, depending on their importance and contribution to the overall performance of the model. The task-specific weights \(\alpha_i\) can be dynamically adjusted during training. Dynamic weighting approaches adjust the weights based on the training dynamics, potentially leading to better overall performance by automatically balancing the learning rates across tasks.
            The optimization of the model parameters \(\theta\) is performed using gradient descent techniques, with backpropagation used to compute the gradients of the loss function with respect to each parameter.
    
    \section{Stable Knowledge Extraction Module} 
        The Top-K choice and memory slot write operations are influenced by the input data's statistical distribution \cite{yang2022inducing, shwartz2024simplifying, dang2024neural}. KEM is no exception.
        To eliminate this issue, we provide an experiment in \ref{sec:skem:toy_exp} and propose stable KEM in \ref{sec:skem:skem_nc}.
    
        \subsection{KEM under Different Distributions}\label{sec:skem:toy_exp}
            The capacity of $R$ is smaller than that of $F$. Assuming the features input to the KEM are balanced, a reasonable knowledge extraction function can be learned using the stochastic gradient descent method. However, if the input features are unbalanced, the KEM's knowledge extraction function will favor the more statistically prevalent features. For example, in vision tasks, the features input to the KEM come from an image, where the background typically occupies half of the features. 
            However, KEM is used as a plugin in MTL. Evaluating KEM with different distributions of \( F \) is quite difficult, as \( P_F \) lacks interpretability and is challenging to construct manually.
            Therefore, we design a toy experiment to demonstrate this claim.

            We conducted toy experiments on Sort-of-clever\cite{santoro2017simple}. Sort-of-CLEVER is a dataset used for visual reasoning and learning. Due to its small size and random generation, it is very suitable for simulating the problem of imbalance in computer vision.
            We have constructed a long-tailed distribution with a power-law exponent of 2, where baseline is a balanced distribution.
            
            \begin{table}
                \centering
                \caption{Comparing sKEM and KEM on different distributions. We show that KEM is sensitive to changes in input distribution by testing it on a long-tailed distribution.}
                \label{tab:nc}
                \begin{tabular}{>{\centering\arraybackslash}m{1cm} >{\centering\arraybackslash}m{2cm} >{\centering\arraybackslash}m{2cm} >{\centering\arraybackslash}m{2cm} >{\centering\arraybackslash}m{2cm} >{\centering\arraybackslash}m{2cm}}
                \toprule
                    & Methods & Use NC & Balance & Imbalance & Accuracy \\
                \midrule
                    1 & KEM & $\times$ & \checkmark & $\times$ & 81\% \\
                    2 & sKEM & \checkmark & \checkmark & $\times$ & 79\% \\
                    3 & KEM & $\times$ & $\times$ & \checkmark & 71\% \\
                    4 & sKEM & \checkmark & $\times$ & \checkmark & 79\% \\
                \bottomrule
                \end{tabular}
            \end{table}
            
            As shown in Table~\ref{tab:nc}, we evaluate KEM on both balanced and imbalanced distributions. The accuracy of KEM drops from 81\% to 71\%, see the first and the third rows of Table~\ref{tab:nc}, indicating its sensitivity to imbalanced inputs. This experiment aims to demonstrate KEM's sensitivity to input variability. In MTL, KEM's input is unknown, underscoring the need for a stable KEM. 
        
        \subsection{Stable KEM with Neural Collapse }\label{sec:skem:skem_nc}

            Inspired by Neural Collapse, we present a straightforward yet effective approach. By projecting the input features into an ETF space before they are processed by the KEM, we improve feature differentiation. This improvement helps the KEM focus less on statistically dominant features and more on diverse attributes, reducing bias~\cite{yang2022inducing, shwartz2024simplifying}.
            To implement this projection, we use a fixed matrix \( W^* \), defined as:
            \begin{equation}
                W^* = \frac{K}{K-1} U \left(I_K - \frac{1}{K} \mathbf{1}_K \mathbf{1}_K^T\right),
            \end{equation}
            where \( W^*, U \in \mathbb{R}^{d \times K} \), and \( U^T U = I_K \), with \( I_K \) representing the identity matrix. Here, \( \mathbf{1}_K \) is an all-ones vector. \( K \) is the number of ETF vertices corresponding to the number of tasks, and \( d \) is the dimension, matching the dimension of KEM.
            The matrix \( W^* \) is then used in the preprocessing step as described in Equation~\ref{eq:step1}. It is applied in a dot product operation with the pre-filtered data. Afterward, the data is filtered through a topk-softmax function and multiplied by \( FW^{vr} \) to complete the attention mechanism, with the results stored in \( \hat{R} \):
            \begin{equation}
                \hat{R} = \text{topk-softmax}\left(\frac{(RW^{qr})(FW^{kr})^T}{\sqrt{d_e}}W^*\right)FW^{vr}.
            \end{equation}

            As shown in Table~\ref{tab:nc}, we evaluate sKEM on balanced and imbalanced distributions. 
            Referring to the first and the third rows of Table~\ref{tab:nc}, the accuracy of sKEM is virtually unchanged, indicating its robustness to imbalanced inputs.
            Referring to the third and the fourth rows of Table~\ref{tab:nc}, the accuracy rose from 71\% to 79\%, indicating sKEM's stability.
            Please note that apart from considering a stable KEM as sKEM in this section, the subsequent section will use KEM to represent sKEM.

    \section{Experiments}
        In this section, we present our experimental setup and evaluation metrics for assessing the performance of KEM across multiple tasks.

        \subsection{Basic Setting}
            \textbf{Datasets.}
                We evaluate KEM with two datasets NYUDv2 \cite{silberman2012indoor} and PASCAL VOC \cite{everingham2010pascal}.
                The NYUDv2 dataset is used for both semantic segmentation and depth estimation tasks. It contains a total of 1449 images, split into 795 images for training and 654 images for validation. For the PASCAL VOC dataset, we utilize the PASCAL-Context split \cite{chen2014detect}, which includes annotations for semantic segmentation, human part segmentation, and composite saliency labels \cite{maninis2019attentive} derived from state-of-the-art pre-trained models \cite{bansal2017pixelnet, chen2018encoder}. This dataset comprises 10,103 images, with 4998 images designated for training and 5105 for validation.

            \textbf{Baselines.}
                In this paper, we compare several MTL methods to evaluate KEM's performance. The methods we selected include MTfomer \cite{xu2022mtformer}, MTL-A \cite{liu2019end}, Cross-stitch \cite{misra2016cross}, MTI-Net \cite{vandenhende2020mti}, Switching \cite{sun2021task}, ERC \cite{bruggemann2021exploring}, NDDR-CNN \cite{gao2019nddr}, PAD-Net \cite{xu2018pad}, Repara \cite{kanakis2020reparameterizing}, AST \cite{maninis2019attentive}, and Auto \cite{bruggemann2020automated}. We use the same dataset and evaluation metrics to ensure a fair comparison.

            \textbf{Evaluation metric.}
                For evaluating our models, we use different metrics based on the specific tasks. Semantic segmentation, saliency estimation, and human part segmentation are evaluated using the mean Intersection over Union (mIoU). The mIoU measures the overlap between the predicted segmentation and the ground truth, averaged over all classes:
                \[ \text{mIoU} = \frac{1}{N} \sum_{i=1}^{N} \frac{\text{TP}_i}{\text{TP}_i + \text{FP}_i + \text{FN}_i} \]
                where \( N \) is the number of classes, \( \text{TP}_i \), \( \text{FP}_i \), and \( \text{FN}_i \) are the true positive, false positive, and false negative pixel counts for class \( i \).
                Depth estimation is evaluated using the Root Mean Square Error (RMSE), which measures the difference between the predicted depth and the actual depth values, providing an indication of the model's accuracy:
                \[ \text{RMSE} = \sqrt{\frac{1}{n} \sum_{i=1}^{n} (p_i - a_i)^2} \]
                where \( n \) is the number of depth values, \( p_i \) is the predicted depth, and \( a_i \) is the actual depth.
                In addition to individual task metrics, we assess the overall performance of MTL approach using a composite metric, $\Delta_m$, adapted from \cite{xu2022mtformer}. This metric is defined as:
                \[
                    \Delta_m = \frac{1}{n} \sum_{i=1}^n (-1)^{l_i} \left( \frac{M_{m,i} - M_{s,i}}{M_{s,i}} \right)
                \]
                where $M_{m,i}$ is the performance of the MTL model on the $i$-th task, $M_{s,i}$ is the performance of the single-task learning (STL) model on the same task, and $l_i$ is 1 if a lower value indicates better performance for the $i$-th task, and 0 otherwise. This composite metric helps us quantify the benefit of multitask learning by comparing it against single-task learning across multiple tasks.

            \textbf{Other implementation details.}
                We employed the AdamW optimizer, which effectively handles sparse gradients and high-dimensional data while also helping to mitigate overfitting. For the PASCAL dataset, we set the learning rate to 0.000025, whereas for the NYUD-v2 dataset, the learning rate was set to 0.00005. The difference in learning rates is adjusted based on the characteristics and requirements of each dataset, ensuring optimal performance during training. The weight decay coefficient for both datasets was set to 0.0001 to further control model complexity and prevent overfitting.
                Additionally, we adopted a polynomial learning rate scheduling strategy. This strategy maintains a higher learning rate at the beginning of training and gradually decreases it over time, ensuring that the model continuously adjusts during convergence to achieve higher accuracy. The pretrained model we used is from Xu\cite{xu2022mtformer}.
                For task configuration, we set the loss weights for semantic segmentation, human part segmentation, and saliency detection in the PASCAL dataset to 1.0, 2.0, and 30.0, respectively. In the NYUD-v2 dataset, the loss weights for semantic segmentation and depth estimation were both set to 1.0. These weight settings are adjusted based on the importance and difficulty of each task.

        \subsection{Main Result of KEM}
\begin{figure}
                \centering
                \includegraphics[width=1.0\textwidth]{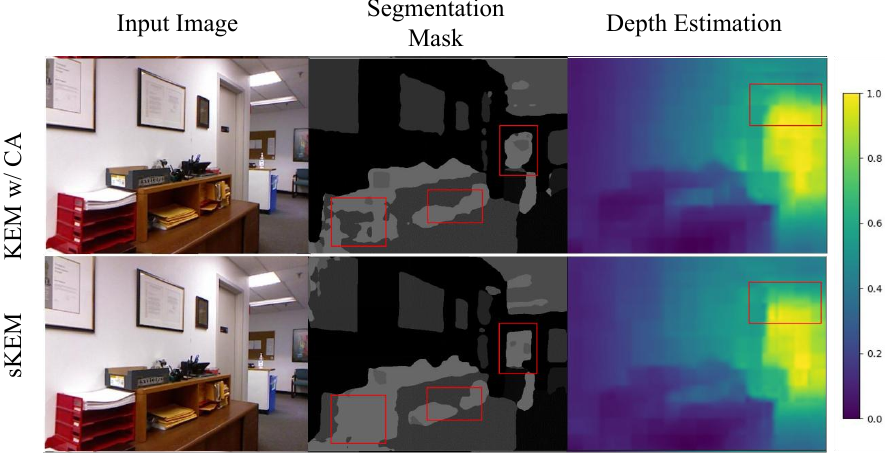}
                \caption{
                    NYUD-v2 validation results on semantic segmentation and depth estimation. Red boxes highlight regions of interest, showing the effectiveness of our method and the baseline with cross-attention.
                }
                \label{fig:compare1}
            \end{figure}

\begin{figure}
                \centering
                \includegraphics[width=1.0\textwidth]{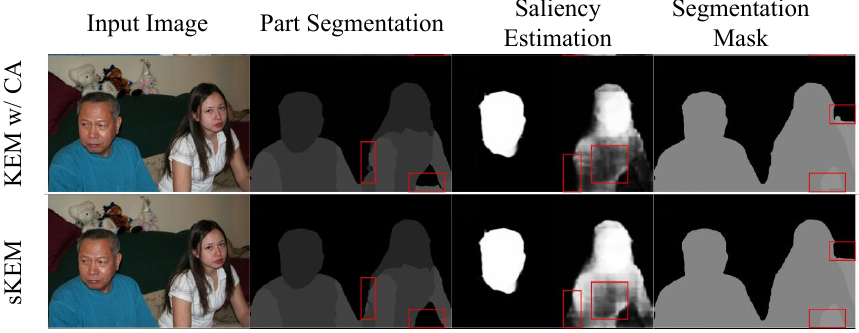}
                \caption{
                    PASCAL validation results on human part segmentation, saliency estimation, and semantic segmentation. Red boxes highlight regions of interest, demonstrating the effectiveness of our method and baseline with cross-attention. In blurry areas of the human body, KEM shows superior noise resistance.
                }
                \label{fig:compare2}
            \end{figure}

            Table~\ref{tab:comparison_methods} presents the experimental results for the NYDU-v2 and PASCAL datasets across all architectures. Our KEM method surpasses most baselines in all tasks. Notably, KEM exhibits exceptional performance in the composite metric, $\Delta_m$. Furthermore, when evaluating the metrics for each individual task, KEM achieves three best and two second-best results.
            \begin{table}
                \centering
                \caption{Comparison on NYUD-v2 and PASCAL Datasets. It can be seen that KEM outperforms the baseline on most metrics.}
                \label{tab:comparison_methods}
                \begin{tabular}{ >{\arraybackslash}m{2cm} >{\centering\arraybackslash}m{1.3cm} >{\centering\arraybackslash}m{1.3cm} >{\centering\arraybackslash}m{1.3cm} >{\centering\arraybackslash}m{1.3cm} >{\centering\arraybackslash}m{1.3cm} >{\centering\arraybackslash}m{1.3cm} >{\centering\arraybackslash}m{1.3cm}}
                    \toprule
                        & \multicolumn{3}{c}{NYUD-v2} & \multicolumn{4}{c}{PASCAL} \\
                        \cmidrule(lr){2-4} \cmidrule(lr){5-8}
                        Method & Seg$\uparrow$ & Dep$\downarrow$ & $\Delta_m$$\uparrow$ & Seg$\uparrow$ & Part$\uparrow$ & Sal$\uparrow$ & $\Delta_m$$\uparrow$ \\
                    \midrule
                        AST&          42.16&          0.570&          -13.39&         68.00&          61.12&          66.10&           -3.24\\
                        Auto&         41.10&          0.541&          -11.58&         64.07&          58.60&          64.92&           -6.99\\
                        Cross-stitch& 41.01&          0.538&          -11.37&         63.28&          60.21&          65.13&           -6.41\\
                        NDDR-CNN&     40.88&          0.536&          -11.30&         63.22&          56.12&          65.16&           -8.56\\
                        MTL-A&        42.03&          0.519&          -8.40&          61.55&          58.89&          64.96&           -7.99\\
                        Repara&       43.22&          0.521&          -7.36&          56.63&          55.85&          59.32&          -14.70\\
                        PAD-Net&      \textbf{50.20}& 0.582&          -6.22&          60.12&          60.70&          67.20&           -6.60\\
                        MTFormer-T &  50.04&          0.490&          +2.87&          73.52&          64.26&          67.24&           +1.55\\
                        Switching&    45.90&          0.527&          -5.17&          64.20&          55.03&          63.31&           -9.59\\
                        MTI-Net&      49.00&          0.529&          -2.14&          64.98&          62.90&          \textbf{67.84}&  -2.86\\      
                        Ours(KEM)&   49.63&          \textbf{0.474}& \textbf{+3.78}& \textbf{73.60}& \textbf{64.94}& 67.82&           \textbf{+2.24}\\
                    \bottomrule
                \end{tabular}
            \end{table}
            Additionally, we conducted an ablation study, as shown in Table~\ref{tab:ab:ca}. We established a baseline (KEM w/ CA) by replacing the IB memory slots in KEM with cross-attention. This allowed us to focus on the core module of our method, minimizing the influence of other training optimization strategies on the contribution of KEM.
            Figures~\ref{fig:compare1} and \ref{fig:compare2} provide a visual comparison between the cross-attention baseline and KEM. It is evident that our method significantly improves noise handling. For instance, in Figure~\ref{fig:compare1}, the first marked location on the segmentation mask clearly shows that KEM demonstrates higher robustness to noise. In Figure~\ref{fig:compare2}, within the blurry areas of the human body, KEM possesses better noise-resistant capabilities.

            \begin{table}
                \centering
                \caption{Ablation study for the soft IB module, replaced with cross-attention. This allowed us to focus on our core method and minimize the impact of other training optimizations on KEM's contribution.}
                \label{tab:ab:ca}
                \begin{tabular}{ >{\arraybackslash}m{2cm} >{\centering\arraybackslash}m{1.3cm} >{\centering\arraybackslash}m{1.3cm} >{\centering\arraybackslash}m{1.3cm} >{\centering\arraybackslash}m{1.3cm} >{\centering\arraybackslash}m{1.3cm} >{\centering\arraybackslash}m{1.3cm} >{\centering\arraybackslash}m{1.3cm}}
                    \toprule
                        & \multicolumn{3}{c}{NYUD-v2} & \multicolumn{4}{c}{PASCAL} \\
                        \cmidrule(lr){2-4} \cmidrule(lr){5-8}
                        Method & Seg$\uparrow$ & Dep$\downarrow$ & $\Delta_m$$\uparrow$ & Seg$\uparrow$ & Part$\uparrow$ & Sal$\uparrow$ & $\Delta_m$$\uparrow$ \\
                    \midrule
                        KEM w/ CA&   48.72&          0.472&          +2.97&          73.45&          64.93&          67.71&           +2.10\\           
                        Ours(KEM)&   \textbf{49.63}&  \textbf{0.474}& \textbf{+3.78}& \textbf{73.60}& \textbf{64.94}& 67.82&           \textbf{+2.24}\\
                    \bottomrule
                \end{tabular}
            \end{table}

            \begin{table}
                \caption{Performance metrics with multiple random seeds showing robust improvements of KEM over baseline.}
                \label{tb:robust-improvements}
                \centering
                \begin{tabular}{>{\centering\arraybackslash}m{1.0cm} >{\centering\arraybackslash}m{1.5cm} >{\centering\arraybackslash}m{1.5cm} >{\centering\arraybackslash}m{1.5cm} >{\centering\arraybackslash}m{1.0cm} >{\centering\arraybackslash}m{1.5cm} >{\centering\arraybackslash}m{1.5cm} >{\centering\arraybackslash}m{1.5cm}}
                \toprule
                    Number & {Seg$\uparrow$} & {Dep$\downarrow$} & {$\Delta_m$$\uparrow$} & Number & {Seg$\uparrow$} & {Dep$\downarrow$} & {$\Delta_m$$\uparrow$} \\
                \midrule
                    0 & 49.48 & 0.477 & +3.55 & 5 & 49.70 & 0.480 & +3.48 \\
                    1 & 49.59 & 0.480 & +3.42 & 6 & 49.35 & 0.474 & +3.72 \\
                    2 & 49.50 & 0.479 & +3.39 & 7 & 49.77 & 0.479 & +3.67 \\
                    3 & 49.30 & 0.477 & +3.36 & 8 & 48.95 & 0.478 & +2.88 \\
                    4 & 49.63 & 0.476 & +3.78 & 9 & 49.31 & 0.479 & +3.17 \\
                \bottomrule
                \end{tabular}
            \end{table}
            \begin{table}
                \caption{Grid search results for memory slot $L$, where $L=20$ represents the optimal configuration identified through our exploration.}
                \label{tb:gs:L}
                \centering
                \begin{tabular}{>{\centering\arraybackslash}m{1.0cm} >{\centering\arraybackslash}m{1.5cm} >{\centering\arraybackslash}m{1.5cm} >{\centering\arraybackslash}m{1.5cm} >{\centering\arraybackslash}m{1.0cm} >{\centering\arraybackslash}m{1.5cm} >{\centering\arraybackslash}m{1.5cm} >{\centering\arraybackslash}m{1.5cm}}
                \toprule
                    $L$ & {Seg$\uparrow$} & {Dep$\downarrow$} & {$\Delta_m$$\uparrow$} & $L$ & {Seg$\uparrow$} & {Dep$\downarrow$} & {$\Delta_m$$\uparrow$} \\
                \midrule
                    5  & 48.92 & 0.477 & 2.98 & 20 & \textbf{49.63} & \textbf{0.474} & \textbf{3.78} \\
                    10 & 49.55 & 0.478 & 3.51 & 25 & 49.44 & 0.476 & 3.57 \\
                    15 & 48.85 & 0.478 & 2.79 & 30 & 49.49 & 0.479 & 3.38 \\ 
                \bottomrule
                \end{tabular}
            \end{table}
            \begin{table}[htbp]
                \caption{Computation and Memory Cost.}
                \centering
                \small
                \begin{tabular}{c|ccc}
                \toprule
                    Method &  KEM & sKEM & Cross-Attention\\
                \midrule
                    Computation (Step/Second) & 10.22 & 7.53 & 3.68 \\
                    GPU Memory (MB) & 265 & 285 & 314 \\
                \bottomrule
                \end{tabular}
                \label{tb:cost}
            \end{table}
            To address concerns regarding the variability of experimental results, we employed multiple random seeds in our experiments. The findings, as detailed in Table~\ref{tb:robust-improvements}, indicate that despite inherent fluctuations, the enhancements achieved by our KEM method over the baseline approaches remain consistently robust and reliable.

        \subsection{Discussion of Expeiments}
            This section will discuss two important hyperparameters: the number of memory slots \( L \) and \( K \) in topk-softmax. 
            The former influences the shared memory space size of the KEM module, while the latter affects the selection number in the Top-K operation during the screening process. Both of these hyperparameters have a significant impact on the KEM module. 
            They cannot be set too high or too low, as this would significantly impact the KEM module.
            In this paper, we employ a grid search method to identify a suitable set of hyperparameters.

            \textbf{Memory Slot $L$.}
                The number of memory slots \( L \) directly determines the size of the memory space, thereby impacting the results of knowledge extraction. A value of \( L \) that is too low cannot accommodate sufficient knowledge, while an excessively high \( L \) increases memory overhead and complexity, resulting in the dilution of the selected knowledge within the memory. To identify the optimal \( L \), we conducted search experiments. As shown in Table~\ref{tb:gs:L}, the experiments on three tasks in the NYUD dataset indicate that the best training results were achieved with \( L \) set to 20. Consequently, other experiments primarily utilized the configuration of \( L \) = 20.

            \textbf{Computation and Memory Cost.}
                
                We have thoroughly compared the computational efficiency and memory usage of KEM, sKEM, and Cross-Attention, as detailed in the Table~\ref{tb:cost}. The results are consistent with our expectations: KEM is the fastest and has the fewest parameters due to its use of Information Bottleneck. sKEM incurs higher costs and more parameters due to the additional projection layer, while Cross-Attention has the lowest efficiency and the highest parameter count.

            \textbf{Top-K operation $K$.}
                $K$ is a crucial parameter in the Top-K operation of the Retrieve step in the knowledge extraction module training. It directly affects the number of items retained during the filtering operation and is essential for managing the extraction process. As $K$ too low can result in excessive filtering, where a significant amount of useful knowledge is not retained for further learning. Conversely, an excessively high $K$ can lead to low filtering effectiveness, retaining too much irrelevant noise information, which causes interference between tasks. Therefore, we conducted an experiment on the $K$ choice.

                The experimental results are recorded in Table~\ref{tb:gs:K}. As can be seen, a comprehensive comparison of the results for the three tasks in the NYUD dataset reveals that the highest training effectiveness of the knowledge extraction module was achieved when the $K$ was set to 3. Consequently, this study primarily used the hyperparameter configuration of $K$=3 for all other experiments.

                \begin{table}
                    \caption{Grid search results for Top-K $K$, where $K=3$ represents the optimal configuration identified through our exploration.}
                    \label{tb:gs:K}
                    \centering
                    \begin{tabular}{>{\centering\arraybackslash}m{1.0cm} >{\centering\arraybackslash}m{1.5cm} >{\centering\arraybackslash}m{1.5cm} >{\centering\arraybackslash}m{1.5cm} >{\centering\arraybackslash}m{1.0cm} >{\centering\arraybackslash}m{1.5cm} >{\centering\arraybackslash}m{1.5cm} >{\centering\arraybackslash}m{1.5cm}}
                    \toprule
                        $K$ & {Seg$\uparrow$} & {Dep$\downarrow$} & {$\Delta_m$$\uparrow$} & $K$ & {Seg$\uparrow$} & {Dep$\downarrow$} & {$\Delta_m$$\uparrow$} \\
                    \midrule
                        1 & 49.42 &0.480 & 3.22 & 4 & 49.47 & 0.477 & 3.50  \\
                        2 & 49.14 &0.476 & 3.26 & 5 & 49.63 & \textbf{0.474} & 3.78 \\
                        3 & \textbf{49.79} &0.476 & \textbf{3.97} & 6 & 49.33 &0.476 & 3.45  \\
                    \bottomrule
                    \end{tabular}
                \end{table}
            
        \subsection{Limitation and future work}
            This study concentrated on computer vision datasets, which often face imbalances and data interference issues. The experiments were limited to this domain, necessitating further validation of multitask learning in other fields.
            KEM's goal is to reduce obvious noise and interference between tasks rather than pinpoint effective features. We aim to propose a framework that identifies effective features and promotes them with learnable weights, which could enhance MTL and lead to a more interpretable model. While we currently lack visualization methods for this process, addressing this challenge is a priority for future exploration.
            Additionally, applying soft IB (information bottleneck) extends beyond MTL; it may also benefit complex single-task learning (STL), such as in NLP with numerous tokens or in 3D reconstruction with many light rays. These areas represent promising avenues for our future research.

    \section{Conclusion}
        In this paper, we propose a novel multi-task learning (MTL) method called KEM. We first re-examine previous Transformer-based cross-attention MTL methods from the perspective of noise. To address the noise issue, we introduce a new method based on the soft information bottleneck (soft IB). Extensive experiments demonstrate that KEM effectively mitigates the interference between tasks. Moreover, we find that KEM significantly reduces computational complexity, which is crucial for efficient resource utilization.
        Additionally, we utilized Neural Collapse to enhance the stability of KEM and proposed sKEM. We have experimentally proven the stability of sKEM.
        In future work, we plan to propose a framework to identify effective features in MTL and extend the model to an interpretable general model. Additionally, we will explore the potential of KEM in single-task scenarios within other complex environments.
    
    \section{Acknowledgments}
        This work was supported by the National Key Research and Development Project of China (2021ZD0110505), the Zhejiang Provincial Key Research and Development Project (2023C01043), and Academy Of Social Governance Zhejiang University.

    \clearpage
    \bibliographystyle{splncs04}
    \bibliography{main}

\begin{thebibliography}{10}
\providecommand{\url}[1]{\texttt{#1}}
\providecommand{\urlprefix}{URL }
\providecommand{\doi}[1]{https://doi.org/#1}

\bibitem{bansal2017pixelnet}
Bansal, A., Chen, X., Russell, B., Gupta, A., Ramanan, D.: Pixelnet: Representation of the pixels, by the pixels, and for the pixels. arXiv preprint arXiv:1702.06506  (2017)

\bibitem{bhattacharjee2022mult}
Bhattacharjee, D., Zhang, T., S{\"u}sstrunk, S., Salzmann, M.: Mult: An end-to-end multitask learning transformer. In: Proceedings of the IEEE/CVF Conference on Computer Vision and Pattern Recognition. pp. 12031--12041 (2022)

\bibitem{bruggemann2020automated}
Bruggemann, D., Kanakis, M., Georgoulis, S., Van~Gool, L.: Automated search for resource-efficient branched multi-task networks. arXiv preprint arXiv:2008.10292  (2020)

\bibitem{bruggemann2021exploring}
Br{\"u}ggemann, D., Kanakis, M., Obukhov, A., Georgoulis, S., Van~Gool, L.: Exploring relational context for multi-task dense prediction. In: Proceedings of the IEEE/CVF international conference on computer vision. pp. 15869--15878 (2021)

\bibitem{chen2023minigpt}
Chen, J., Zhu, D., Shen, X., Li, X., Liu, Z., Zhang, P., Krishnamoorthi, R., Chandra, V., Xiong, Y., Elhoseiny, M.: Minigpt-v2: large language model as a unified interface for vision-language multi-task learning. arXiv preprint arXiv:2310.09478  (2023)

\bibitem{chen2018encoder}
Chen, L.C., Zhu, Y., Papandreou, G., Schroff, F., Adam, H.: Encoder-decoder with atrous separable convolution for semantic image segmentation. In: Proceedings of the European conference on computer vision (ECCV). pp. 801--818 (2018)

\bibitem{chen2014detect}
Chen, X., Mottaghi, R., Liu, X., Fidler, S., Urtasun, R., Yuille, A.: Detect what you can: Detecting and representing objects using holistic models and body parts. In: Proceedings of the IEEE conference on computer vision and pattern recognition. pp. 1971--1978 (2014)

\bibitem{dang2024neural}
Dang, H., Tran, T., Nguyen, T., Ho, N.: Neural collapse for cross-entropy class-imbalanced learning with unconstrained relu feature model. arXiv preprint arXiv:2401.02058  (2024)

\bibitem{dittadi2020transfer}
Dittadi, A., Tr{\"a}uble, F., Locatello, F., W{\"u}thrich, M., Agrawal, V., Winther, O., Bauer, S., Sch{\"o}lkopf, B.: On the transfer of disentangled representations in realistic settings. arXiv preprint arXiv:2010.14407  (2020)

\bibitem{everingham2010pascal}
Everingham, M., Van~Gool, L., Williams, C.K., Winn, J., Zisserman, A.: The pascal visual object classes (voc) challenge. International journal of computer vision  \textbf{88},  303--338 (2010)

\bibitem{fang2021exploring}
Fang, C., He, H., Long, Q., Su, W.J.: Exploring deep neural networks via layer-peeled model: Minority collapse in imbalanced training. Proceedings of the National Academy of Sciences  \textbf{118}(43),  e2103091118 (2021)

\bibitem{gao2019nddr}
Gao, Y., Ma, J., Zhao, M., Liu, W., Yuille, A.L.: Nddr-cnn: Layerwise feature fusing in multi-task cnns by neural discriminative dimensionality reduction. In: Proceedings of the IEEE/CVF conference on computer vision and pattern recognition. pp. 3205--3214 (2019)

\bibitem{goyal2021coordination}
Goyal, A., Didolkar, A., Lamb, A., Badola, K., Ke, N.R., Rahaman, N., Binas, J., Blundell, C., Mozer, M., Bengio, Y.: Coordination among neural modules through a shared global workspace. arXiv preprint arXiv:2103.01197  (2021)

\bibitem{han2022survey}
Han, K., Wang, Y., Chen, H., Chen, X., Guo, J., Liu, Z., Tang, Y., Xiao, A., Xu, C., Xu, Y., et~al.: A survey on vision transformer. IEEE transactions on pattern analysis and machine intelligence  \textbf{45}(1),  87--110 (2022)

\bibitem{hong2024concept}
Hong, J., Park, K.H., Pavlic, T.P.: Concept-centric transformers: Enhancing model interpretability through object-centric concept learning within a shared global workspace. In: Proceedings of the IEEE/CVF Winter Conference on Applications of Computer Vision. pp. 4880--4891 (2024)

\bibitem{hu2024revisiting}
Hu, Y., Xian, R., Wu, Q., Fan, Q., Yin, L., Zhao, H.: Revisiting scalarization in multi-task learning: A theoretical perspective. Advances in Neural Information Processing Systems  \textbf{36} (2024)

\bibitem{im2017denoising}
Im~Im, D., Ahn, S., Memisevic, R., Bengio, Y.: Denoising criterion for variational auto-encoding framework. In: Proceedings of the AAAI conference on artificial intelligence. vol.~31 (2017)

\bibitem{ji2021unconstrained}
Ji, W., Lu, Y., Zhang, Y., Deng, Z., Su, W.J.: An unconstrained layer-peeled perspective on neural collapse. arXiv preprint arXiv:2110.02796  (2021)

\bibitem{kanakis2020reparameterizing}
Kanakis, M., Bruggemann, D., Saha, S., Georgoulis, S., Obukhov, A., Van~Gool, L.: Reparameterizing convolutions for incremental multi-task learning without task interference. In: Computer Vision--ECCV 2020: 16th European Conference, Glasgow, UK, August 23--28, 2020, Proceedings, Part XX 16. pp. 689--707. Springer (2020)

\bibitem{lachapelle2023synergies}
Lachapelle, S., Deleu, T., Mahajan, D., Mitliagkas, I., Bengio, Y., Lacoste-Julien, S., Bertrand, Q.: Synergies between disentanglement and sparsity: Generalization and identifiability in multi-task learning. In: International Conference on Machine Learning. pp. 18171--18206. PMLR (2023)

\bibitem{liu2022stateful}
Liu, D., Shah, V., Boussif, O., Meo, C., Goyal, A., Shu, T., Mozer, M., Heess, N., Bengio, Y.: Stateful active facilitator: Coordination and environmental heterogeneity in cooperative multi-agent reinforcement learning. arXiv preprint arXiv:2210.03022  (2022)

\bibitem{liu2023deep}
Liu, J., Hao, J., Lin, H., Pan, W., Yang, J., Feng, Y., Wang, G., Li, J., Jin, Z., Zhao, Z., et~al.: Deep learning-enabled 3d multimodal fusion of cone-beam ct and intraoral mesh scans for clinically applicable tooth-bone reconstruction. Patterns  \textbf{4}(9) (2023)

\bibitem{liu2023parameter}
Liu, J., Hu, T., Zhang, Y., Feng, Y., Hao, J., Lv, J., Liu, Z.: Parameter-efficient transfer learning for medical visual question answering. IEEE Transactions on Emerging Topics in Computational Intelligence  (2023)

\bibitem{liu2019end}
Liu, S., Johns, E., Davison, A.J.: End-to-end multi-task learning with attention. In: Proceedings of the IEEE/CVF conference on computer vision and pattern recognition. pp. 1871--1880 (2019)

\bibitem{maninis2019attentive}
Maninis, K.K., Radosavovic, I., Kokkinos, I.: Attentive single-tasking of multiple tasks. In: Proceedings of the IEEE/CVF conference on computer vision and pattern recognition. pp. 1851--1860 (2019)

\bibitem{menon2020pulse}
Menon, S., Damian, A., Hu, S., Ravi, N., Rudin, C.: Pulse: Self-supervised photo upsampling via latent space exploration of generative models. In: Proceedings of the ieee/cvf conference on computer vision and pattern recognition. pp. 2437--2445 (2020)

\bibitem{miladinovic2019disentangled}
Miladinovi{\'c}, {\DJ}., Gondal, M.W., Sch{\"o}lkopf, B., Buhmann, J.M., Bauer, S.: Disentangled state space representations. arXiv preprint arXiv:1906.03255  (2019)

\bibitem{misra2016cross}
Misra, I., Shrivastava, A., Gupta, A., Hebert, M.: Cross-stitch networks for multi-task learning. In: Proceedings of the IEEE conference on computer vision and pattern recognition. pp. 3994--4003 (2016)

\bibitem{montero2020role}
Montero, M.L., Ludwig, C.J., Costa, R.P., Malhotra, G., Bowers, J.: The role of disentanglement in generalisation. In: International Conference on Learning Representations (2020)

\bibitem{muhammad2020deep}
Muhammad, K., Ullah, A., Lloret, J., Del~Ser, J., de~Albuquerque, V.H.C.: Deep learning for safe autonomous driving: Current challenges and future directions. IEEE Transactions on Intelligent Transportation Systems  \textbf{22}(7),  4316--4336 (2020)

\bibitem{papyan2020prevalence}
Papyan, V., Han, X., Donoho, D.L.: Prevalence of neural collapse during the terminal phase of deep learning training. Proceedings of the National Academy of Sciences  \textbf{117}(40),  24652--24663 (2020)

\bibitem{razavi2019generating}
Razavi, A., Van~den Oord, A., Vinyals, O.: Generating diverse high-fidelity images with vq-vae-2. Advances in neural information processing systems  \textbf{32} (2019)

\bibitem{santoro2017simple}
Santoro, A., Raposo, D., Barrett, D.G., Malinowski, M., Pascanu, R., Battaglia, P., Lillicrap, T.: A simple neural network module for relational reasoning. Advances in neural information processing systems  \textbf{30} (2017)

\bibitem{shwartz2024simplifying}
Shwartz-Ziv, R., Goldblum, M., Li, Y., Bruss, C.B., Wilson, A.G.: Simplifying neural network training under class imbalance. Advances in Neural Information Processing Systems  \textbf{36} (2024)

\bibitem{silberman2012indoor}
Silberman, N., Hoiem, D., Kohli, P., Fergus, R.: Indoor segmentation and support inference from rgbd images. In: Computer Vision--ECCV 2012: 12th European Conference on Computer Vision, Florence, Italy, October 7-13, 2012, Proceedings, Part V 12. pp. 746--760. Springer (2012)

\bibitem{sun2021task}
Sun, G., Probst, T., Paudel, D.P., Popovi{\'c}, N., Kanakis, M., Patel, J., Dai, D., Van~Gool, L.: Task switching network for multi-task learning. In: Proceedings of the IEEE/CVF international conference on computer vision. pp. 8291--8300 (2021)

\bibitem{tirer2022extended}
Tirer, T., Bruna, J.: Extended unconstrained features model for exploring deep neural collapse. In: International Conference on Machine Learning. pp. 21478--21505. PMLR (2022)

\bibitem{tirer2023perturbation}
Tirer, T., Huang, H., Niles-Weed, J.: Perturbation analysis of neural collapse. In: International Conference on Machine Learning. pp. 34301--34329. PMLR (2023)

\bibitem{tucker2021emergent}
Tucker, M., Li, H., Agrawal, S., Hughes, D., Sycara, K., Lewis, M., Shah, J.A.: Emergent discrete communication in semantic spaces. Advances in Neural Information Processing Systems  \textbf{34},  10574--10586 (2021)

\bibitem{van2017neural}
Van Den~Oord, A., Vinyals, O., et~al.: Neural discrete representation learning. Advances in neural information processing systems  \textbf{30} (2017)

\bibitem{vandenhende2021multi}
Vandenhende, S., Georgoulis, S., Van~Gansbeke, W., Proesmans, M., Dai, D., Van~Gool, L.: Multi-task learning for dense prediction tasks: A survey. IEEE transactions on pattern analysis and machine intelligence  \textbf{44}(7),  3614--3633 (2021)

\bibitem{vandenhende2020mti}
Vandenhende, S., Georgoulis, S., Van~Gool, L.: Mti-net: Multi-scale task interaction networks for multi-task learning. In: Computer Vision--ECCV 2020: 16th European Conference, Glasgow, UK, August 23--28, 2020, Proceedings, Part IV 16. pp. 527--543. Springer (2020)

\bibitem{wang2020graph}
Wang, W., Xu, H., Gan, Z., Li, B., Wang, G., Chen, L., Yang, Q., Wang, W., Carin, L.: Graph-driven generative models for heterogeneous multi-task learning. In: Proceedings of the AAAI Conference on Artificial Intelligence. vol.~34, pp. 979--988 (2020)

\bibitem{wang2024balance}
Wang, Y., Li, L., Yang, J., Lin, Z., Wang, Y.: Balance, imbalance, and rebalance: Understanding robust overfitting from a minimax game perspective. Advances in neural information processing systems  \textbf{36} (2024)

\bibitem{xin2024mmap}
Xin, Y., Du, J., Wang, Q., Yan, K., Ding, S.: Mmap: Multi-modal alignment prompt for cross-domain multi-task learning. In: Proceedings of the AAAI Conference on Artificial Intelligence. vol.~38, pp. 16076--16084 (2024)

\bibitem{xu2018pad}
Xu, D., Ouyang, W., Wang, X., Sebe, N.: Pad-net: Multi-tasks guided prediction-and-distillation network for simultaneous depth estimation and scene parsing. In: Proceedings of the IEEE conference on computer vision and pattern recognition. pp. 675--684 (2018)

\bibitem{xu2022mtformer}
Xu, X., Zhao, H., Vineet, V., Lim, S.N., Torralba, A.: Mtformer: Multi-task learning via transformer and cross-task reasoning. In: European Conference on Computer Vision. pp. 304--321. Springer (2022)

\bibitem{yang2022inducing}
Yang, Y., Chen, S., Li, X., Xie, L., Lin, Z., Tao, D.: Inducing neural collapse in imbalanced learning: Do we really need a learnable classifier at the end of deep neural network? Advances in neural information processing systems  \textbf{35},  37991--38002 (2022)

\bibitem{zhang2024scalable}
Zhang, R., Liu, J., Li, Z., Dong, H., Fu, J., Wu, C.: Scalable geometric fracture assembly via co-creation space among assemblers. In: Proceedings of the AAAI Conference on Artificial Intelligence. vol.~38, pp. 7269--7277 (2024)

\bibitem{zhang2021survey}
Zhang, Y., Yang, Q.: A survey on multi-task learning. IEEE Transactions on Knowledge and Data Engineering  \textbf{34}(12),  5586--5609 (2021)

\bibitem{zhong2023understanding}
Zhong, Z., Cui, J., Yang, Y., Wu, X., Qi, X., Zhang, X., Jia, J.: Understanding imbalanced semantic segmentation through neural collapse. In: Proceedings of the IEEE/CVF Conference on Computer Vision and Pattern Recognition. pp. 19550--19560 (2023)

\bibitem{zhu2021geometric}
Zhu, Z., Ding, T., Zhou, J., Li, X., You, C., Sulam, J., Qu, Q.: A geometric analysis of neural collapse with unconstrained features. Advances in Neural Information Processing Systems  \textbf{34},  29820--29834 (2021)

\end{thebibliography}
\end{document}